\documentclass[10pt,twocolumn,letterpaper]{article}

\usepackage{cvpr}              

\usepackage{colortbl}
\usepackage{amssymb}
\usepackage{pifont}
\usepackage{booktabs,multirow}
\usepackage{makecell}
\usepackage{tabulary}
\usepackage{fontawesome5}
\usepackage{bbding}

\newlength\savewidth

\usepackage{amsmath}
\usepackage{capt-of}
\usepackage[dvipsnames]{xcolor}
\usepackage{xspace}
\usepackage{enumitem}
\usepackage{amsmath,amssymb,amsbsy,amsfonts,dsfont,pifont,bm,bbm,mathrsfs,mathtools,nicefrac}
\usepackage{algorithm,algpseudocode,listings}
\usepackage{booktabs,multirow,adjustbox,diagbox,threeparttable,tabularray}

\newcommand{\methodname}{SynMotion}
\newcommand{\method}{\texttt{\methodname}\xspace}

\newcommand{\tocite}[1]{{\color{red} [TO CITE]}}

\definecolor{CQColor}{rgb}{0.0,0.0,1.0} 

\definecolor{TSColor}{rgb}{0.5,0.0,0.8} 

\definecolor{CQRColor}{rgb}{1.0,0.0,1.0} 

\definecolor{cvprblue}{rgb}{0.21,0.49,0.74}
\usepackage[pagebackref,breaklinks,colorlinks,allcolors=cvprblue]{hyperref}
\usepackage{wrapfig}
\usepackage[capitalize]{cleveref}  
\crefname{section}{Sec.}{Secs.}
\Crefname{section}{Section}{Sections}
\crefname{table}{Tab.}{Tabs.}
\Crefname{table}{Table}{Tables}
\crefname{figure}{Fig.}{Figs.}
\Crefname{figure}{Figure}{Figures}
\crefname{equation}{Eq.}{Eqs.}
\Crefname{equation}{Equation}{Equations}
\definecolor{baseColor}{rgb}{0.75,0.05,0.1}

\definecolor{checkmarkColor}{rgb}{0.1,0.75,0.1}

\definecolor{demphcolor}{RGB}{144,144,144}

\def\paperID{1231} 
\def\confName{CVPR}
\def\confYear{2026}

\title{\methodname: Semantic-Visual Adaptation for Motion \\ Customized Video Generation}

\author{Shuai Tan$^{1,2}$\footnotemark[1]\>\,, Biao Gong$^{2}$\footnotemark[2]\>\,\footnotemark[3]\>\,, Yujie Wei$^{3}$, Shiwei Zhang$^{3}$, Zhuoxin Liu$^{4}$, Ke Ma$^{5}$,\\ 
Yan Wang$^{6}$, Kecheng Zheng$^{2}$, Xing Zhu$^{2}$, Yujun Shen$^{2}$, Hengshuang Zhao$^{1}$\footnotemark[3]\>\,\\[5pt]
{$^1$The University of Hong Kong\ \ $^2$Ant Group\ \ $^3$Tongyi\ \  $^4$University of Wisconsin–Madison} \\
{$^5$ Huazhong University of Science and Technology\ \ $^6$University of North Carolina at Chapel Hill} \\
}

\begin{document}

\twocolumn[{
\maketitle
\begin{center}
    \vspace{-2pt}
    \includegraphics[width=0.95\linewidth]{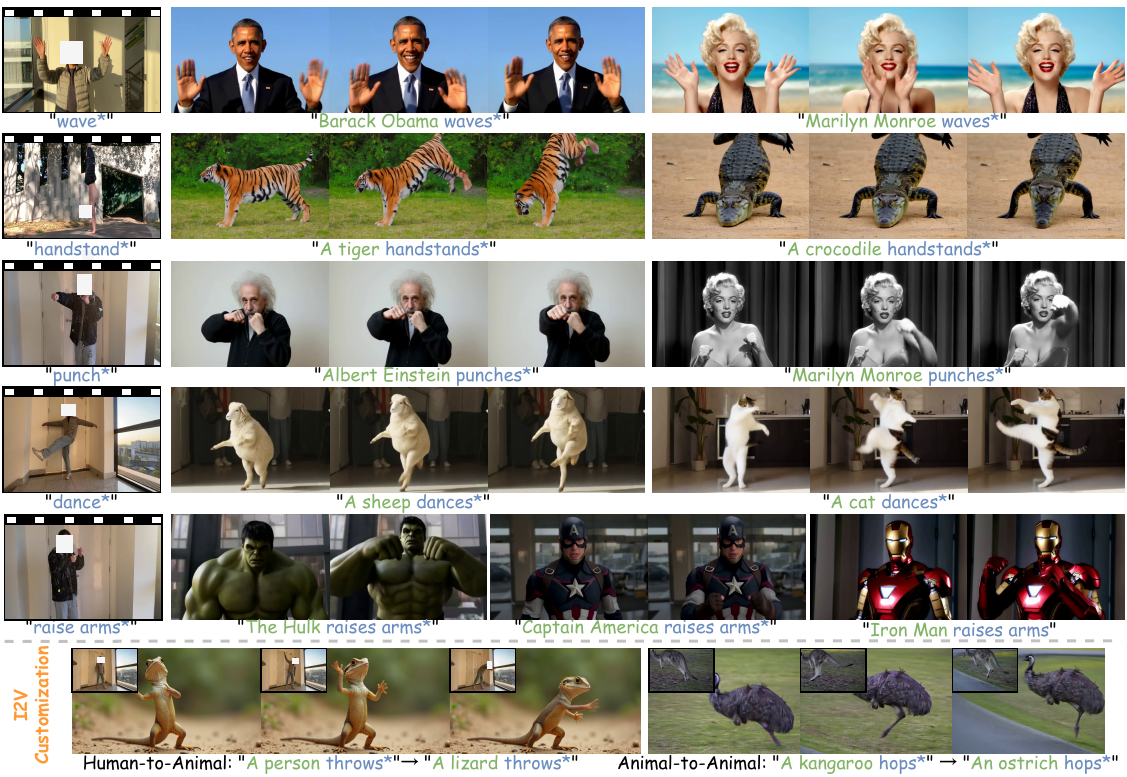}
        \vspace{-6pt}
    \captionof{figure}{
        \textbf{Generated customization results of our \method.} Given a few exemplar videos demonstrating a common motion, \method learns motion pattern and successfully synthesizes diverse subjects performing the same action in both T2V and I2V setting.
    }
    \vspace{2mm}
    \label{fig:teaser}
\end{center}
}]

\footnotetext[1]{Work done during internship at Ant Group. \footnotemark[2]Project lead.}
\footnotetext[3]{Corresponding author.}

\begin{abstract}
Diffusion-based video motion customization facilitates the acquisition of human motion from a few video samples, while achieving arbitrary subjects transfer through precise textual conditioning. Existing approaches often rely on semantic-level alignment, expecting the model to learn new motion concepts and combine them with other entities (\textit{e.g.}, ``\textit{cats}'' or ``\textit{dogs}'') to produce visually appealing results. However, video data involve complex spatio-temporal patterns, and focusing solely on semantics cause the model to overlook the visual complexity of motion. Conversely, tuning only the visual representation leads to semantic confusion in representing the intended action. %
To address these limitations, we propose \method, a motion-customized video generation model that jointly leverages semantic guidance and visual adaptation.
\textit{At the semantic level}, we introduce the dual-embedding semantic comprehension mechanism which disentangles subject and motion, allowing the model to learn customized motion features while preserving its generative capabilities for diverse subjects. 
\textit{At the visual level}, we integrate parameter-efficient motion adapters into a pre-trained video generation model to enhance motion fidelity and temporal coherence. 
Furthermore, we introduce a new embedding-specific training strategy which \textbf{alternately optimizes} subject and motion embeddings, supported by the manually constructed Subject Prior Video (SPV) training dataset. This strategy promotes motion specificity while preserving generalization across diverse subjects.
Lastly, we introduce MotionBench, a newly curated benchmark with diverse motion patterns. Experimental results across both T2V and I2V settings show that \method outperforms state-of-the-arts (SOTAs).
Project page: \href{https://lucaria-academy.github.io/SynMotion/}{https://lucaria-academy.github.io/SynMotion/}
\end{abstract}

\section{Introduction}
\label{sec:intro}
Recent advances in diffusion-based video generation models~\cite{make-a-video, modelscopet2v, tft2v, tune-a-video, chai2023stablevideo, ceylan2023pix2video, guo2023animatediff, zhou2022magicvideo, an2023latent} enable powerful text-to-video (T2V) and image-to-video (I2V) generation, allowing the synthesis of realistic
videos, which significantly expands the creative scope of synthetic media. However, the model struggles to learn and generalize certain specialized action semantics or uncommon human actions rarely present in challenging training dataset (\textit{e.g., ``handstands''}), resulting in unnatural videos for various subjects and environments. To address these limitations, recent studies have introduced the task of motion-customized video generation~\cite{wu2024customcrafter, zhang2023motioncrafter, bi2024customttt}, which aims to synthesize videos depicting imaginative scenarios, such as \textit{``a crocodile performs a handstand''} or \textit{``Marilyn Monroe punches''}, by transferring specific motions from user-provided reference videos to diverse target subjects. Nonetheless, developing a robust and versatile motion customization framework that effectively supports both T2V and I2V scenarios remains challenging and has not been sufficiently explored.

Existing approaches to motion customization can be categorized into two types: semantic-level methods~\cite{gal2022textualinversion, ruiz2022dreambooth, kumari2022customdiffusion} and visual-level methods~\cite{meral2024motionflow,yesiltepe2024motionshop, zhang2025training,zhang2025training, jeong2024vmc, zhao2024motiondirector}. Semantic-level approaches, such as ADI~\cite{adi2024gong} and ReVersion~\cite{shi2024relationbooth}, operate by injecting novel concept identifiers into pre-trained T2V models~\cite{stablediffusion, stablediffusion2} to represent motion semantics. However, their direct adaptation from image generation to video synthesis is non-trivial, due to two fundamental challenges: (1) video generation requires enhanced semantic comprehension, particularly for temporally coherent motion understanding\cite{mimir,hunyuanvideo}, and (2) the increased parameter complexity from temporal modeling in video escalates training difficulty and exacerbates frame inconsistency issues\cite{cogvideox}. As shown in Fig.~\ref{fig:compare_inversion}, these gaps make it challenging to extend them to the video domain. In contrast, visual-level approaches, such as Motion Inversion~\cite{zhang2023motioncrafter}, directly optimize motion-specific latent representations within the visual feature space. While effective at motion reproduction, they often capture rigid, instance-specific patterns rather than abstract, transferable motion concepts~\cite{dmt}, limiting generalization to closely related subjects. Moreover, most visual-based methods preserve the original spatial layout of reference video, including object positions and background~\cite{jeong2024vmc}, which further restricts the diversity of the generated content. These limitations indicate that motion customization cannot be adequately addressed by either semantic or visual modeling alone. Instead, a hybrid approach is needed to balance motion expressiveness, subject generalization, and video diversity.

In this paper, we propose \method, a unified framework that combines semantic comprehension and visual adaptation to enable precise and generalizable motion customization video generation.
\textbf{At the semantic level}, \method builds on HunyuanVideo~\cite{hunyuanvideo}, enhanced with decoder-only large language models (LLMs)\cite{glm2024chatglm, 2023xtuner} to provide strong semantic grounding. In addition, we introduce a dual-embedding semantic comprehension mechanism that decomposes LLM-generated embeddings into subject and motion components using prompt-aware partitioning, allowing the model to retain its native subject synthesis capabilities while adapting to new motion patterns. Furthermore, a learnable embedding refiner is designed to fuse subject and motion representations in the latent space, ensuring interaction between them. \textbf{At the visual level}, 
we inject lightweight, trainable motion adapters into the frozen generation model to improve motion realism and temporal coherence. With the combined power of semantic comprehension and visual adaptation, our method exhibits strong generalization across motions and subjects, enabling diverse subjects to perform a wide range of motions.

Furthermore, to prevent interference between subject and motion semantics, we introduce the \textbf{embedding-specific training strategy}. The strategy requires constructing an auxiliary training dataset, which differs from user-provided motion examples that usually consist of a few videos of uncommon motions. Instead, it contains diverse subjects paired videos with common motions, called subject prior videos (SPV), enabling the model to perceive a wider range of content beyond the customized video distribution and achieve better generalization. Specifically, during training, we dynamically alternate between real customization samples and SPV-generated videos, selectively updating the motion and subject embeddings based on their relevance. This training approach encourages the model to learn precise motion embeddings while maintaining strong subject generalization, thereby enabling flexible and high-quality video generation, as shown in Fig.~\ref{fig:teaser}.

Lastly, we introduce a new benchmark called MotionBench, which is designed to evaluate motion customization performance across various motion categories. We conduct comprehensive experiments under both T2V and I2V settings using MotionBench. Quantitative and qualitative comparisons against recent methods demonstrate that \method achieves state-of-the-art (SOTA) results in both motion alignment and video quality, particularly excelling in subject-motion disentanglement and output diversity. The code and weights will soon be open-sourced.
\section{Related Work}
\label{sec:related_work}
\subsection{Video diffusion models}

In recent years, diffusion models~\cite{DDIM, DDPM} have demonstrated remarkable generative capabilities, particularly in the field of video generation. Early video generation approaches~\cite{make-a-video, modelscopet2v, tft2v,tune-a-video,chai2023stablevideo,ceylan2023pix2video,guo2023animatediff,zhou2022magicvideo,an2023latent,xing2023simda,qing2023hierarchical,yuan2023instructvideo, longcatvideo,wei2024dreamvideo,wei2023online, wei2025dreamrelation,wei2025routing} extended pretrained text-to-image models~\cite{Dalle2, saharia2022photorealistic, stablediffusion} with an additional temporal layer to support video synthesis. VDM~\cite{ho2022video} pioneers the use of diffusion models for video generation by directly modeling the pixel-level distribution of videos. Subsequent works such as ModelScopeT2V~\cite{modelscopet2v} and VideoCrafter~\cite{chen2023videocrafter1, chen2024videocrafter2} further advance text-to-video synthesis by incorporating spatiotemporal modules into the generation pipeline. With the emergence of Diffusion Transformers~\cite{peebles2023scalable}, several works~\cite{opensora, opensoraplan, cogvideo, cogvideox, wan2025, genmo2024mochi,ji2025memflow,ji2025physmaster,ji2025layerflow} have leveraged their scalable architecture to achieve more compelling video generation results. However, prior methods often relied on CLIP~\cite{CLIP} or T5~\cite{t5} for semantic encoding, which limited their capacity for text understanding~\cite{xie2024sana}. Mimir~\cite{mimir} firstly introduces a decoder-only large language model (LLM)~\cite{phi3} into video generation, enabling more precise semantic control. Similarly, HunyuanVideo~\cite{hunyuanvideo} incorporates a MLLM~\cite{sun2024hunyuanlargeopensourcemoemodel, glm2024chatglm, 2023xtuner} to
support complex reasoning in video synthesis.

\subsection{Customized video generation}
Customized video generation
aims to synthesize videos based on user-provided concepts, such as subject identity~\cite{yuan2024identity, wu2024motionbooth, zhou2024sugar, wu2024videomaker, zhang2025fantasyid, she2025customvideox, wu2024customcrafter}, motion patterns~\cite{wu2024customcrafter,zhang2023motioncrafter,bi2024customttt,meral2024motionflow}, or relational cues~\cite{shi2024relationbooth,wei2025dreamrelation,huang2024reversion,zhang2024inv,ge2024customizing}. 
Depending on the representation used for customization, existing approaches can be broadly categorized into two main types: semantic-level methods and visual-level methods.

\noindent \textbf{Semantic-level methods.} Semantic-level methods focus on learning latent concepts (e.g., subject identity~\cite{ tan2025edtalk, tan2025edtalk++, tan2024flowvqtalker, tan2024style2talker}, motion~\cite{tan2024animate, tan2025animate,tan2024say, tan2025fixtalk, tan2023emmn,codance}, or relationship~\cite{wei2025dream}) and injecting them into text-conditioned diffusion models to enable controlled generation. A large body of work in this direction is based on diffusion-based inversion, which can be further categorized into optimization-based~\cite{gal2022textualinversion, ruiz2022dreambooth, kumari2022customdiffusion, li2023generate, han2023svdiff, hu2022lora, choi2023custom, kawar2022imagic, voynov2023p+, alaluf2023neural}, encoder-based~\cite{wei2023elite, jia2023tamingencoder, xu2023prompt, zhou2023enhancing, ma2023subject, ye2023ip,miao2025rose}, and hybrid methods~\cite{gal2023encoder, chen2023disenbooth, gong2023talecrafter, arar2023domain, li2023blipdiffusion, ruiz2024hyperdreambooth}. These approaches typically learn one or more subject-specific tokens or embeddings from a few reference images, which are later composed with textual prompts to synthesize personalized content in novel contexts. Among these, ADI~\cite{adi2024gong} is the most relevant to our work. It introduces layer-wise identifier tokens to better encode action semantics from static images, enabling limited motion customization. However, these methods are not directly applicable to video generation tasks, which require richer temporal reasoning and pose challenges in frame consistency and parameter efficiency. In contrast, our method 
targets motion customization in video generation through a novel semantic decomposition scheme.

\noindent \textbf{Visual-level methods.}
Visual-level methods~\cite{zhang2023motioncrafter, bi2024customttt,dmt, wu2024customcrafter, meral2024motionflow, yesiltepe2024motionshop, zhang2025training, jeong2024vmc, kothandaraman2024imposter,li2024spherehead,li2025hyplanehead,li2026condition} approach customization from the video feature space, often transferring motion directly from reference videos. Some methods use latent optimization~\cite{zhang2023motioncrafter, bi2024customttt}, others apply structured constraints~\cite{dmt, wu2024customcrafter, meral2024motionflow, yesiltepe2024motionshop, zhang2025training, jeong2024vmc, kothandaraman2024imposter}, such as space-time loss of DMT~\cite{dmt} or temporal embeddings of Motion Inversion~\cite{mi}. While effective at reproducing visual dynamics, they tend to overfit instance-specific trajectories and retain the original scene layout, limiting subject generalization and output diversity.
In contrast, semantic-level methods offer better generalization through high-level concept control, but lack precision in motion synthesis. Therefore, effective motion-customized video generation requires a joint modeling of both semantic and visual aspects, which is the core insight of our method.

\section{Methodology}
\label{sec:method}
\begin{figure*}[t]
   \begin{center}
    \includegraphics[width=1\linewidth]{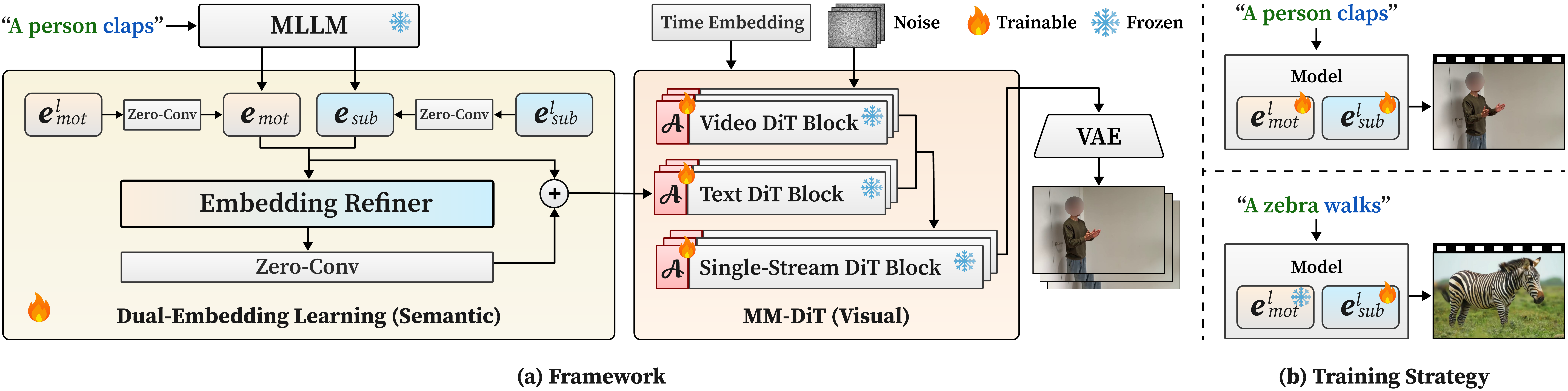}
    \end{center}
    \vspace{-4mm}
    \caption{\textbf{The pipeline of \method.} Given a prompt in the form of $<\text{subject}, \text{motion}>$, we use a MLLM to obtain the corresponding text embedding, which is then decomposed into a subject embedding $e_{sub}$ and motion embedding $e_{mot}$. Each part is augmented with the learnable embeddings (i.e., $e^l_{sub}$ and $e^l_{mot}$) in a zero-initialized convolutional residual (Zero-Conv) $\mathcal{Z}$ manner. These embeddings are then passed through an Embedding Refiner $\mathcal{R}$, which fuses subject and motion semantics. The refined embeddings are reintegrated via Zero-Conv $\mathcal{Z}$ and injected into the video generation backbone. An additional Adapter $\mathcal{R}$ enhances motion-aware features, enabling the final model to generate videos with customized motion across novel subjects.}
    \label{fig:method}
    \vspace{-3mm}
\end{figure*}

In this paper, our goal is to generate videos that demonstrate a particular motion observed in a few exemplar videos, with the main subjects of these videos aligning with textual prompts. We begin by outlining the preliminaries of diffusion models in Sec.~\ref{sec:pre}. Then, we delve into the details of \method in Sec.~\ref{sec:Inverse-X}, which enhances motion customization through semantic and visual level, along with the Embedding-Specific Training Strategy in Sec~\ref{sec:train_strategy}.

\subsection{Preliminary}
\label{sec:pre}

Recent advancements in video generation frequently utilize the MM-DiT architecture, which employs full attention to process both video and text tokens simultaneously. Video tokens are compressed using a 3D causal VAE. Unlike previous methods that rely on the T5 as a text encoder, HunyuanVideo uses a decoder-only LLM and offers detailed instructions for encoding text tokens, enabling more nuanced semantic information. Subsequently, HunyuanVideo incorporates diffusion processes to simulate the reverse of a Markov chain with a length of $T$. At timestep $t$, noise $\epsilon$ is added to $z$ to obtain a noise-corrupted latent $z_t$:
\begin{equation}
\mathcal{L}=\mathbb{E}_{\mathcal{E}(x), \epsilon \in \mathcal{N}(0,1), e_\theta, t}\left[\left\|\epsilon-\epsilon_{\theta}\left(\boldsymbol{z}_{t}, e_\theta, t\right)\right\|_{2}^{2}\right],
\end{equation}
where $\epsilon_{\theta}$ is the MM-DiT and $e_\theta$ refers to the text embedding. After the reversed denoising stage, the predicted clean latent is further used
to reconstruct the predicted video.

\subsection{\method}
\label{sec:Inverse-X}

\noindent \textbf{Dual-Embedding Learning.}
A straightforward idea for motion customization is to directly apply textual inversion~\cite{gal2022textualinversion}, as successfully demonstrated in image generation tasks. However, as shown in Fig.~\ref{fig:compare_inversion}, naively adapting word-level inversion to video generation fails to yield satisfactory results. This is primarily because video generation requires richer semantic comprehension, especially for temporal and motion-related concepts, which cannot be adequately captured by simple token-level embeddings. To address this limitation, we delve into the embedding space and propose a more structured Dual-Embedding Semantic Comprehension that decomposes and refines semantic representations for both subjects and motions control in video synthesis.

As shown in Fig.~\ref{fig:method} (a), given a prompt of the form $<\text{subject}, \text{motion}>$, we first extract its semantic representation using a MLLM, resulting in a single text embedding. We then perform prompt-aware decomposition to split this embedding into two components: subject embedding $e_{sub}$ and motion embedding $e_{mot}$, based on the semantic position and role of tokens in original prompt. To endow these components with learnability while preserving the model’s original understanding of language, we attach learnable residual embeddings (i.e., $e^l_{mot}$ and $e^l_{sub}$) via Zero-Conv layer $\mathcal{Z}$. To better guide the learning of these residuals, we carefully design their initialization strategy. The learnable motion embedding $e^l_{mot}$ is responsible for capturing motion-specific characteristics of the customized motion. To accelerate convergence and facilitate meaningful learning, we initialize it with the original motion embedding (the embedding of the word `\textit{clap}' extracted via the MLLM) derived from the prompt. Considering the causal nature of decoder-only MLLMs, where each token is influenced by preceding tokens, we take the embedding of complete phrase `\textit{a person claps}' as the initialization for motion embedding. In contrast, the learnable subject embedding $e^l_{sub}$ aims to preserve the model’s native generalization across diverse subjects. Thus, we randomly initialize it, allowing the model to freely adapt to novel subject appearances without being biased by prior textual semantics. Next, we introduce an Embedding Refiner $\mathcal{R}$, which facilitates semantic interaction between the subject and motion embeddings. This refiner allows for better alignment between subject appearance and motion semantics in latent space. The refined embeddings are then added back to original embeddings through another Zero-Conv, preserving both the base semantics and learned customization:
\begin{equation}
    e = [e_{mot} + \mathcal{Z}(e^l_{mot}), e_{sub} + \mathcal{Z}(e^l_{sub})], \quad  e' = e+\mathcal{Z}(\mathcal{R}(e))
\end{equation}

\noindent \textbf{Motion-Aware Adapter.}
Finally, the combined embeddings are injected into the frozen pre-trained video generation backbone. However, relying solely on semantic-level customization is insufficient for capturing the dynamic nature of motion in videos. This is primarily due to the increased parameter complexity introduced by temporal modeling in video architectures, which significantly escalates training difficulty and exacerbates frame inconsistency issues. As a result, semantic-only approaches often produce limited motion amplitudes and fail to replicate the dynamic characteristics of videos. 

To address this, we further delve into visual-level modeling. The visual denoising process is implemented via MM-DiT, which consists of Text DiT Blocks, Video DiT Blocks, and Single-Stream DiT Blocks. For each component, we introduce lightweight motion-aware low-rank adaptation modules $\mathcal{A}$,
which further enhances the model’s ability to capture and represent motion, leading to improved temporal consistency and fidelity. 
Specifically, for each attention layer composed of $\{Q,K,V\}$, adapters are incorporated following a low-rank residual way: $\tilde{\mathbf{W}}_* = \mathbf{W}_* + \Delta \mathbf{W}_* = \mathbf{W}_* + \mathbf{B}_* \mathbf{A}_*, * \in \{Q,K,V\}$, where $\mathbf{A}_* \in \mathbb{R}^{r \times d}$ and $\mathbf{B}_* \in \mathbb{R}^{d \times r}$ are learnable matrices with rank $r \ll d$, and $\mathbf{W}_*$ remains frozen during training. In this way, the combination of semantic-level customization
and visual-level enhancement through parameter-efficient adaptation enables our framework to generalize across a wide range of subjects and motions, achieving flexible video generation.

\subsection{Embedding-Specific Training Strategy}
\label{sec:train_strategy}
To ensure that the learnable subject and motion embeddings capture their intended semantics without interfering with each other, we propose an Embedding-Specific Training Strategy. This strategy enables disentangled learning of subject and motion features, and preserves the original diversity and generative capacity of the pre-trained video generation model. Specifically, in addition to the user-provided example videos for a specific motion (\textit{e.g.}, `\textit{clap}'), we construct an auxiliary dataset called Subject Prior Videos (SPV), where we sample a variety of animal categories (\textit{e.g.}, `\textit{cat}', `\textit{zebra}') and pair them with common motions (\textit{e.g}., `\textit{run}', `\textit{walk}') to form generic prompts like ``\textit{a zebra walks}''. These prompts are then passed through the frozen video generation model to synthesize videos that reflect diverse subject-motion combinations
. SPV serves as a subject-centric prior, helping the model maintain its ability to generalize to unseen subjects beyond human figures.

During training, we define a sampling probability $\alpha \in [0,1]$. As shown in Fig.~\ref{fig:method} (b), at each training step, with probability $\alpha$, we sample a real user-provided example video, and jointly optimize both the learnable motion embedding and learnable subject embedding, as both motion and subject are relevant to the customization goal. By contrast, with the probability $1-\alpha$, we sample a subject prior video. Since its motion is not related to the target customized motion, we freeze the learnable motion embedding during this phase, which regularizes the subject embedding to retain generalization across a wide range of entities. By jointly training on example videos and SPVs, the model achieves a balance between precise motion customization and robust subject generalization, enabling flexible video synthesis without sacrificing quality or diversity.

\subsection{MotionBench}
The primary objective of this work is to extract a representative motion from a few exemplar videos in which different subjects perform the same action. In order to provide a standardized setting for systematic comparison on this task, we introduce MotionBench, a new benchmark that covers a diverse set of motion categories. Specifically, we first query GPT-4~\cite{chatgpt,gpt4} to generate 30 candidate motion categories. Each motion is then combined with a subject to form a textual prompt in the format of $<\text{subject}, \text{motion}>$, such as ``\textit{a cat is knocking the door}'', which is subsequently fed into the pre-trained video generation model~\cite{hunyuanvideo}. We retain only those motions for which the model fails to produce satisfactory outputs, ensuring that the selected actions are indeed non-trivial for generation. 
We define a curated set of 26 unique actions, and for each action, we collect 20 real-world exemplar videos to serve as the final benchmark.

\begin{figure*}[t]
  \centering
    \includegraphics[width=1\linewidth]{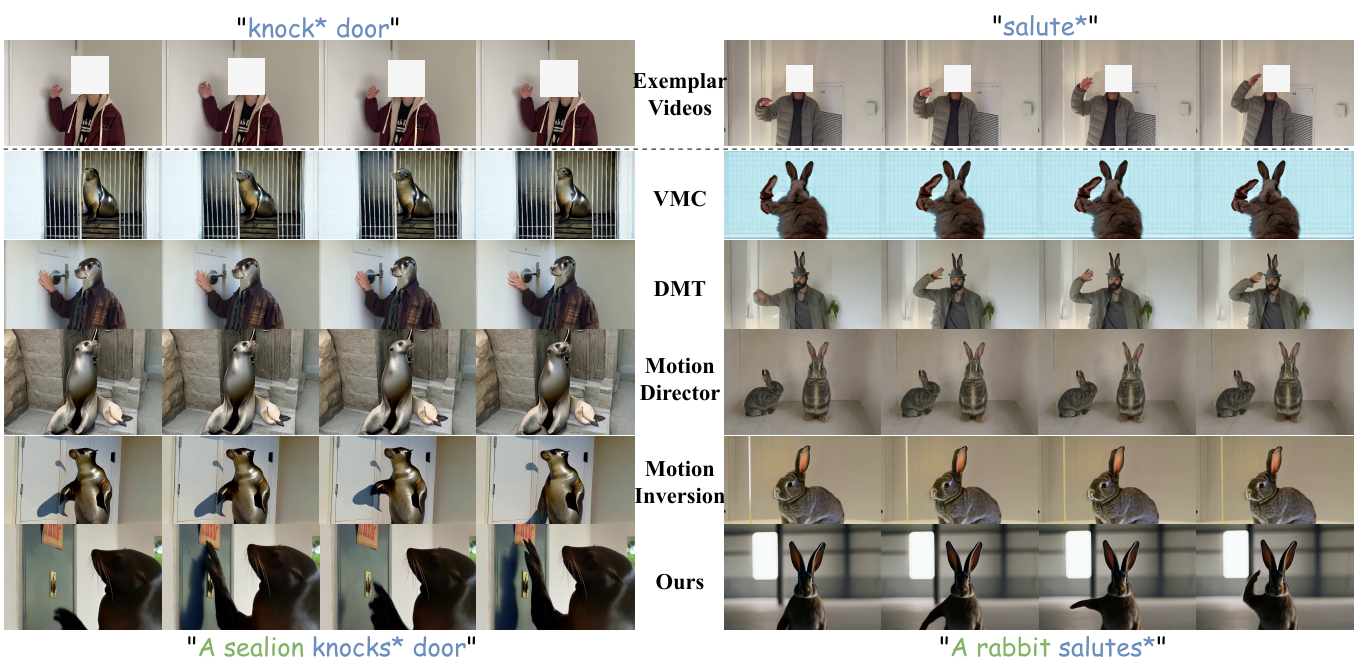}
    \vspace{-5mm}
    \caption{Qualitative comparisons with state-of-the-art motion customization methods.}
    \label{fig:compare_motion}
    \vspace{-15pt}
\end{figure*}

\section{Experiments}
\label{sec:exp}

\subsection{Experiment Setup}
\noindent \textbf{Implementation Details.} 
In addition to the proposed MotionBench, we train our model on the open-sourced FlexiACT dataset~\cite{flexiact}, which contains a wide array of complex real-world motions and various fitness exercises. For each motion category, we train the model for 2,000 iterations using the AdamW optimizer with a learning rate of 2e-5. The sampling probability parameter $\alpha$ is empirically set to 0.75. We adopt HunyuanVideo~\cite{hunyuanvideo} as our base video generation model, conducted on 8 H20 GPUs.

\noindent \textbf{Baselines.} 
We compare our method against two categories of baselines: (a) Visual-level methods, including VMC~\cite{jeong2024vmc}, DMT~\cite{dmt}, Motion Director~\cite{zhao2024motiondirector} and Motion Inversion~\cite{mi}. These methods share the same goal as ours, which learns a specific motion from given exemplar videos and generalizing it to new subjects. (b) Semantic-level methods, including Textual Inversion~\cite{gal2022textualinversion}, DreamBooth~\cite{ruiz2022dreambooth}, and ReVersion~\cite{huang2024reversion}. Since originally proposed for customized image generation, we adapt Textual Inversion and DreamBooth to the motion customization video domain by implementing them on top of HunyuanVideo. For ReVersion, we follow~\cite{wei2025dreamrelation} to implement it using the Mochi~\cite{genmo2024mochi}. We also compare MotionClone~\cite{motionclone}, a training-free framework that enables motion cloning, in the Appendix. All baseline methods are trained and evaluated on our proposed MotionBench for a fair comparison.

\begin{figure*}[t]
  \centering
    \includegraphics[width=1\linewidth]{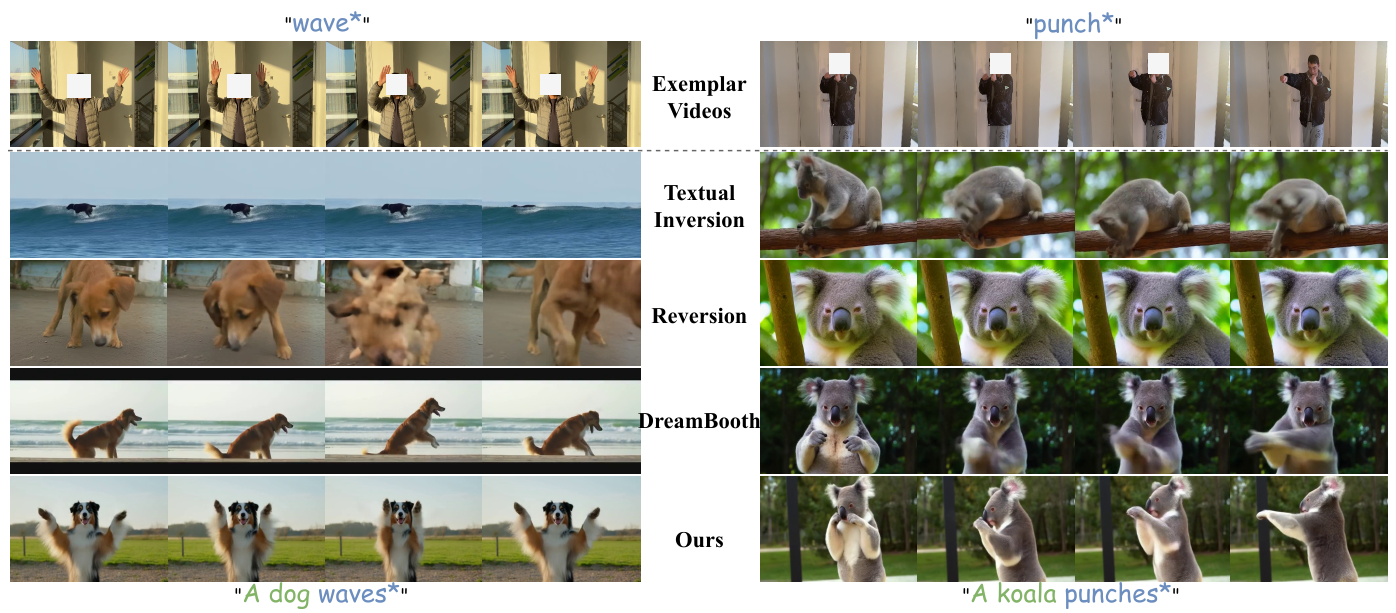}
    \vspace{-3mm}
    \captionsetup{skip=-1pt}
    \caption{Qualitative comparisons with state-of-the-art semantic-level  methods.}
    \label{fig:compare_inversion}
    \vspace{-3mm}
\end{figure*}

\noindent \textbf{Evaluation metrics.} 
We evaluate our method across
: (1) Motion Quality. To assess whether the generated video accurately reflects the intended motion, we leverage SOTA Vision-Language Models (VLMs). Specifically, we input each generated video into QwenVL~\cite{Qwen-VL,Qwen2-VL,Qwen2.5-VL}, a powerful visual question answering (VQA) model, and prompt it with a yes/no question asking whether the video depicts the specified motion. The binary responses are converted into an accuracy score. This process is repeated 10 times for each motion category, and the average accuracy is reported. In addition, to evaluate motion smoothness, an essential aspect of motion quality, we adopt the motion smoothness score from VBench~\cite{vbench}. (2) Subject Quality. Another important factor is whether the generated subject remains consistent with the entity described in the prompt. Similarly, we utilize QwenVL to verify it and compute the subject accuracy based on yes/no responses. In addition, we assess subject consistency using the corresponding metric from VBench, which evaluates whether the subject's appearance remains visually coherent throughout the entire video. (3) Video Quality. We further evaluate the visual quality of generated videos from three complementary perspectives: imaging quality, dynamic degree, and background consistency. All metrics are adopted from VBench~\cite{vbench}, providing a comprehensive assessment of perceptual clarity, temporal stability, and motion richness. To further strengthen our quantitative analysis, we include {FVD} for perceptual quality, {Temporal Consistency} for frame stability, {CLIP-T} for global text-video alignment, and Flow Score for motion preservation.

\subsection{Experiment Results}

\begin{table*}[!t]
\setlength\tabcolsep{6pt}
\def\w{50pt} 
\centering\footnotesize
\begin{tabular}{l@{\extracolsep{4pt}}c@{\extracolsep{4pt}}c@{\extracolsep{4pt}}c@{\extracolsep{4pt}}c@{\extracolsep{4pt}}c@{\extracolsep{4pt}}c@{\extracolsep{4pt}}c@{\extracolsep{4pt}}c@{\extracolsep{4pt}}c@{\extracolsep{4pt}}c@{\extracolsep{4pt}}c}
\hline
\multirow{2}{*}{\textbf{Method}} & \textbf{Motion} & \textbf{Motion} & \textbf{Subject} & \textbf{Subject} & \textbf{Imaging} & \textbf{Dynamic} & \textbf{Background} & \textbf{CLIP} & \textbf{FVD} & \textbf{FVD} & \textbf{Flow} \\
 & \textbf{Accuracy} & \textbf{Consistency} & \textbf{Accuracy} & \textbf{Consistency}& \textbf{Quality}& \textbf{Degree} & \textbf{Consistency} & \textbf{T} & \textbf{3DRN50} & \textbf{3DInception} & \textbf{Score} \\
\hline
VMC~\cite{jeong2024vmc} & 53.64\% & 98.70\% & 38.43\% & 95.97\% & 58.70\% & 20.60\% & 95.89\% & 0.293 & 395.32 & 5242.49 & 0.67 $\pm$ 0.12 \\
DMT~\cite{dmt} & 51.16\% & 98.68\% & 34.88\% & 95.43\% & 58.28\% & 12.50\% & 95.96\% & 0.291  & 390.06 & 4546.08 & 0.69 $\pm$ 0.07 \\
MotionDirector~\cite{zhao2024motiondirector} & 41.67\% & 99.01\% & 71.93\% & 97.56\% & 59.44\% & 3.51\% & 96.85\% & 0.299  & 465.60 & 6485.23 & 0.80 $\pm$ 3.13 \\
MotionInversion~\cite{mi} & 59.31\% & 98.98\% & 73.21\% & 96.74\% & 59.23\% & 3.57\% & 96.65\% & 0.295 & 213.04 & 3361.62 & 0.97 $\pm$ 0.36 \\
Textual Inversion~\cite{gal2022textualinversion} & 21.43\% & 98.91\% & 62.94\% & 95.98\% & 65.85\% & 47.06\% & 96.02\% & 0.277 & 456.23 & 4614.82 & 1.77 $\pm$ 0.67 \\
DreamBooth~\cite{ruiz2022dreambooth} & 37.56\% & 98.68\% & 69.76\% & 93.60\% & 66.66\% & 69.77\% & 94.84\% & 0.278 & 385.82 & 3500.53 & 1.92 $\pm$ 1.09 \\
\hline
\textbf{\method} & \textbf{68.60\%} & \textbf{99.50\%} & \textbf{97.67\%} & \textbf{98.26\%} & \textbf{69.47\%} & \textbf{88.24\%} & \textbf{97.59\%} & \textbf{0.322 } & \textbf{212.05} & \textbf{3129.19} & \textbf{0.41 $\pm$ 0.02} \\
\hline
\end{tabular}
\vspace{-2mm}
\caption{%
    Quantitative comparison results. The best results for each column are \textbf{bold}.
}
  \label{tab:comparison}%
\end{table*}%

\noindent \textbf{Qualitative Comparison}
Fig.~\ref{fig:compare_motion} presents a qualitative comparison between our method and existing visual-level motion customization baselines. VMC is able to generate subjects mentioned in the prompt (\textit{e.g.}, `\textit{rabbit}', `\textit{sea lion}', `\textit{door}'), but fails to produce the correct motion. While DMT shows improved consistency with the exemplar motion, it rigidly preserves the scene layout and enforces strict motion alignment with the exemplar video, which limits the diversity of generated content. This constraint also restricts its generalization to semantically distant subjects. For example, when the target subject differs significantly from the exemplar (\textit{e.g.}, generating a rabbit from a human exemplar), DMT produces anatomically inconsistent results, such as human arms on a rabbit. Motion Director produces duplicate subjects (\textit{e.g.}, two rabbits). Although Motion Inversion improves subject identity and realism, it still fails to reproduce the correct motion. In contrast, our method accurately captures the motion in the exemplar videos and generalizes it to semantically distant subjects such as rabbits and sea lions. It preserves subject (\textit{e.g.}, generating proper flippers for a sea lion) while maintaining temporal coherence and generating intended motion.

\begin{figure}[t]
   \begin{center}
\includegraphics[width=1.0\linewidth]{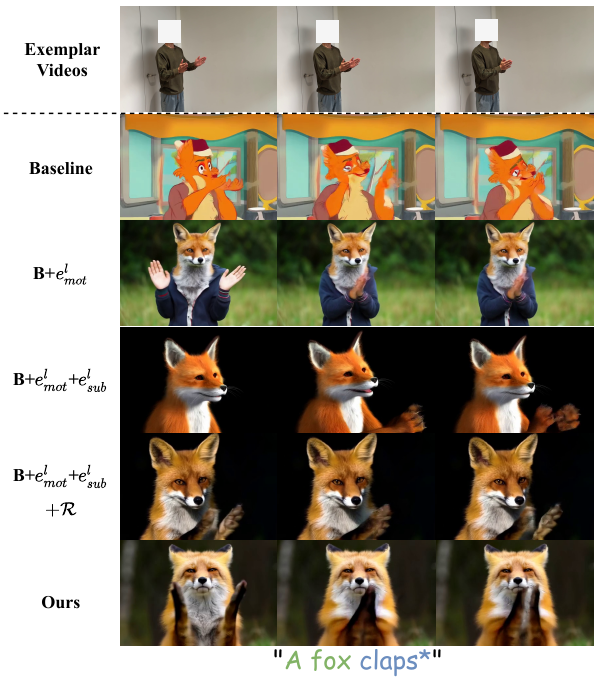}
\end{center}
    \vspace{-6mm}
     \caption{
        Visualization of ablation study.
    }
	\label{fig:ablation_study}
    \vspace{-4mm}
\end{figure}

We further compare our approach with semantic-level methods, as shown in Fig.~\ref{fig:compare_inversion}. These methods benefit from the ability to encode the motion observed in the exemplar video into a newly learned token, which can then be flexibly composed into new prompts for video generation. As a result, compared to motion customization baselines, they tend to preserve greater diversity in the output, including variations in background and scene layout. Specifically, Textual Inversion attempts to map the entire motion into a single learned token. However, due to the inherent complexity of motion and the challenges of video generation, it fails to capture the correct action. Similarly, ReVersion leverages relation-steering contrastive learning to steer motion prompts based on linguistic priors, but it still struggles to reproduce the motion observed in the exemplar videos. DreamBooth, which fine-tunes the video generation model's parameters, achieves modest improvements in cases like `\textit{punch}', but the results remain blurry and suffer from poor temporal consistency. In more complex cases, such as `\textit{wave}', DreamBooth fails to learn the intended motion altogether. In contrast, our method accurately captures the target motion from exemplar videos and generates visually clear, temporally coherent sequences across various subjects. These comparisons further demonstrate the effectiveness and superiority of our approach in motion-specific video generation tasks.

\begin{figure*}[t]
  \centering
  \includegraphics[width=1\linewidth]{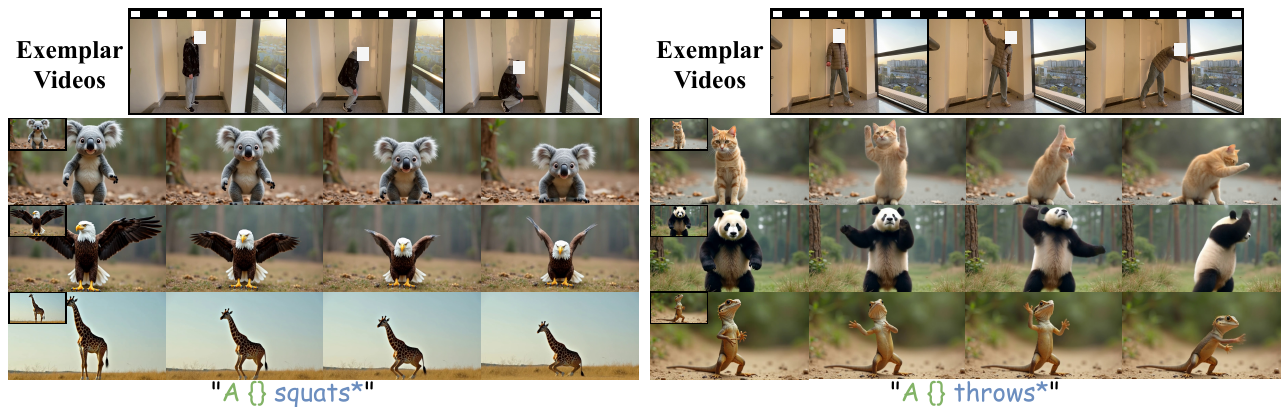}
  \captionsetup{skip=-1pt}
    \caption{The results of our method generalized on Image-to-Video task. 
    } 
    \label{fig:i2v_result}
\end{figure*}

\noindent \textbf{Quantitative Comparison}
Tab.~\ref{tab:comparison} presents quantitative comparison between our method and several SOTA baselines. 
VMC~\cite{jeong2024vmc} and DMT~\cite{dmt} preserve the exemplar structure (\textit{e.g.}, spatial layout, and even subject morphology), leading to high motion accuracy but low subject accuracy and dynamic degree.
MotionDirector~\cite{zhao2024motiondirector} improves subject alignment but sacrifices motion accuracy. MotionInversion~\cite{mi} balances motion and subject alignment, yet lacks diversity due to its rigid motion representation.
In contrast, semantic-level methods~\cite{gal2022textualinversion, ruiz2022dreambooth} allow flexible recombination of learned motion tokens with arbitrary prompts, resulting in higher dynamic degree and output diversity. Nonetheless, they fail to capture accurate motions from exemplars, reflected by lower motion accuracy scores. Our proposed method successfully addresses these limitations by accurately modeling the desired motion, generating semantically correct subjects, and maintaining temporal and structural diversity. Consequently, it achieves the best overall performance.

\noindent \textbf{User Study}
We conducted a user study to evaluate our proposed method, involving 20 volunteers who rated 20 video groups generated by five different methods. Each group consisted of the five generated videos alongside a textual prompt and a reference video. Evaluations were based on user ratings across three key aspects: Motion Alignment, Subject Alignment, and overall Video Quality. The results in Tab.~\ref{tab:user_study} clearly indicate that our method was most preferred by users across all three criteria.

\begin{table}[t]
  \centering
  \resizebox{\linewidth}{!}{
  \begin{tabular}{@{}l@{\hspace{2pt}}|@{\hspace{2pt}}c@{\hspace{2pt}}@{\hspace{2pt}}c@{\hspace{2pt}}@{\hspace{2pt}}c@{}} 
    \hline
    Metric/Method & \textbf{Motion Alignment} & \textbf{Subject Alignment} & \textbf{Video Quality} \\
    \hline
DMT  & 8.1\% $\pm$ 5.57 & 5.0\% $\pm$ 2.56 & 3.9\% $\pm$ 1.28  \\
MotionInversion  &7.2\% $\pm$ 2.80 & 5.6\% $\pm$ 1.10 & 4.4\% $\pm$ 1.88\\
MotionDirector  & 3.1\% $\pm$ 0.68 & 3.3\% $\pm$ 1.00 & 3.3\% $\pm$ 2.0 \\
MotionClone  & 3.3\% $\pm$ 1.44 & 4.2\% $\pm$ 4.14 & 4.4\% $\pm$ 2.43 \\
\hline
\textbf{\method} & \textbf{78.3\%} $\pm$ 1.34 &  \textbf{81.9\%} $\pm$ 2.41 & \textbf{83.9\%} $\pm$ 3.25  \\

    \hline
  \end{tabular}
  }
  \vspace{-2mm}
  \caption{User study results.}
  \label{tab:user_study}
  \vspace{-3mm}
\end{table}

\noindent \textbf{Ablation Study}
To assess the effectiveness of each component in \method, we conduct a progressive ablation study by incrementally introducing our designed modules on top of a baseline. Specifically, we set HunyuanVideo as our Baseline and add $e^l_{mot}$, $e^l_{sub}$, $\mathcal{R}$ and $\mathcal{A}$  step by step. As illustrated in Fig.~\ref{fig:ablation_study}, the baseline produces a cartoon-style anthropomorphic fox, failing to capture the intended motion. Upon introducing $e^l_{mot}$, the model generates the correct action, but human-like hands appear on the fox. This issue is resolved after incorporating $e^l_{sub}$, which restores a subject-appropriate appearance, although the overall visual quality remains unnatural. Next, $\mathcal{R}$ enables effective semantic fusion, producing more coherent and contextually aligned results. Nevertheless, the motion magnitude still falls short of matching the exemplar videos. Finally, by integrating $\mathcal{A}$, our model is able to generate natural and semantically faithful videos that align well with both the subject and the desired motion. Moreover, we report quantitative results in Appendix, which further clarify the individual contributions of each component.

\subsection{Generalization on I2V}

To further evaluate the generalization capability of our approach, we integrate the proposed Dual-Embedding Learning and Motion-Aware Adapter into an image-to-video (I2V) generation framework. Specifically, we implement our method on top of HunyuanVideo-I2V and conduct experiments on our proposed MotionBench. As shown in Fig.~\ref{fig:i2v_result}, our method successfully enables the subject in the input image to perform the customized motion specified by the exemplar video. These results demonstrate that our framework is not only effective in text-to-video (T2V) scenarios, but also generalizes well to the I2V setting, validating its robustness and versatility across different input modalities.

\section{Conclusion}

In this paper, we propose \method that learns motion patterns and generalizes them to diverse subjects from both semantic and visual perspectives. For the former, we introduce a dual-embedding mechanism that captures customized motion features while maintaining the model’s generative flexibility across various subjects. For the latter, we incorporate a motion adapter into a pre-trained video generation model to enhance motion fidelity and ensure temporal coherence. To further improve motion specificity without sacrificing diversity or generalization, we adopt an embedding-specific training strategy that facilitates robust embedding learning. To support systematic evaluation, we propose MotionBench, a benchmark encompassing diverse and challenging motion categories. Extensive experiments demonstrate that \method outperforms SOTAs, achieving superior performance.

\section*{Acknowledgments}
This work is supported by the Hong Kong Research Grant Council General Research Fund (No. 17213925) and National Natural Science Foundation of China (No. 62422606, 62441615).
\break
{
    \small
    \bibliographystyle{ieeenat_fullname}
    \bibliography{main}
}


\end{document}